\documentclass[10pt,twocolumn,letterpaper]{article}

\usepackage[pagenumbers]{cvpr} % To force page numbers, e.g. for an arXiv version

\usepackage[dvipsnames]{xcolor}
\usepackage{bm}
\usepackage{amsmath}
\usepackage{comment}
\usepackage{float}

\definecolor{cvprblue}{rgb}{0.21,0.49,0.74}
\usepackage[pagebackref,breaklinks,colorlinks,citecolor=cvprblue]{hyperref}

\def\paperID{11656} % *** Enter the Paper ID here
\def\confName{CVPR}
\def\confYear{2024}

\title{X-Adapter: Adding Universal Compatibility of Plugins for Upgraded Diffusion Model}

\author{Lingmin Ran$^1$ \and Xiaodong Cun$^3$ \and Jia-Wei Liu$^1$ \and Rui Zhao$^1$ \and Song Zijie$^4$ \and Xintao Wang$^3$ \and Jussi Keppo$^2$ \and Mike Zheng Shou$^{1,}$\footnotemark[1]
\and \\ $^1$Show Lab, \,\,$^2$National University of Singapore \and \\$^3$Tencent AI Lab \and \\ $^4$Fudan University}

\begin{document}

% \maketitle
\twocolumn[{
\maketitle
\begin{center}
    \captionsetup{type=figure}
    \vspace{-2em}
    \includegraphics[width=1.\textwidth]{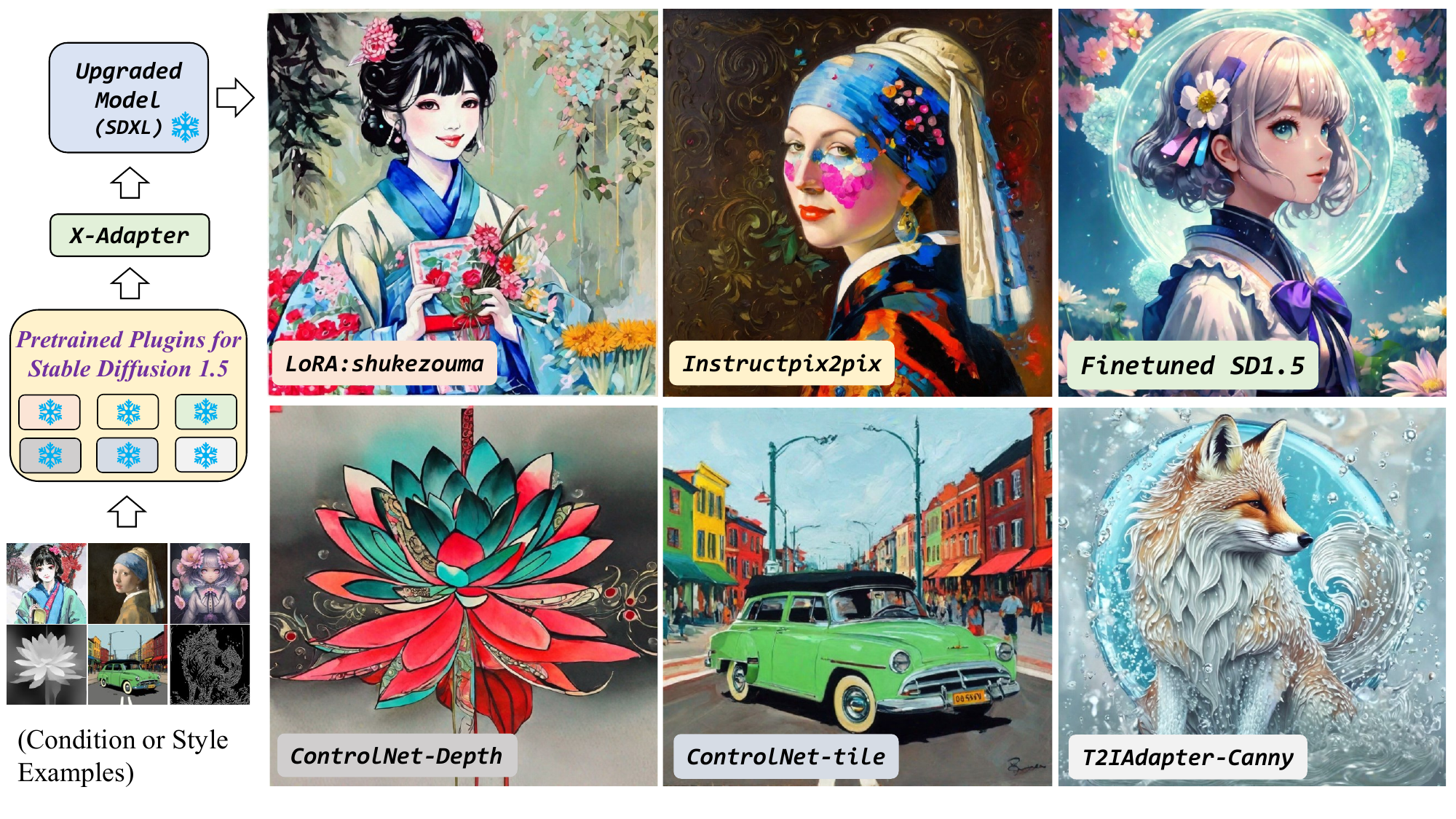}
    \vspace{-2em}
    \captionof{figure}{ 
    Given pretrained plug-and-play modules~(\eg, ControlNet, LoRA) of the base diffusion model~(\eg, Stable Diffusion 1.5), the proposed X-Adapter can universally upgrade these plugins, enabling them directly work with the upgraded Model~(\eg, SDXL) without further retraining. Text prompts: \textit{``1girl, solo, smile, looking at viewer, holding flowers"} \textit{``Apply face paint"} \textit{``1girl, upper body, flowers"} \textit{``A colorful lotus, ink"} \textit{``Best quality, extremely detailed"} \textit{``A fox made of water"} from left to right, top to bottom. 
    }
\end{center}
}]

\renewcommand{\thefootnote}{\fnsymbol{footnote}}
\footnotetext[1]{~Corresponding Author.}

\begin{abstract}
\vspace{-1em}
We introduce X-Adapter, a universal upgrader to enable the pretrained plug-and-play modules~(\eg, ControlNet, LoRA) to work directly with the upgraded text-to-image diffusion model~(\eg, SDXL) without further retraining. 
We achieve this goal by training an additional network to control the frozen upgraded model with the new text-image data pairs. 
In detail, X-Adapter keeps a frozen copy of the old model to preserve the connectors of different plugins. Additionally,  X-Adapter adds trainable mapping layers that bridge the decoders from models of different versions for feature remapping. The remapped features will be used as guidance for the upgraded model. To enhance the guidance ability of X-Adapter, we employ a null-text training strategy for the upgraded model.
After training, we also introduce a two-stage denoising strategy to align the initial latents of X-Adapter and the upgraded model. 
Thanks to our strategies, X-Adapter demonstrates universal compatibility with various plugins and also enables plugins of different versions to work together, thereby expanding the functionalities of diffusion community. To verify the effectiveness of the proposed method, we conduct extensive experiments and the results show that X-Adapter may facilitate wider application in the upgraded foundational diffusion model. Project page at: \url{https://showlab.github.io/X-Adapter/}. 
\end{abstract}   
\vspace{-1em}
\section{Introduction}
\label{sec:intro}

% finetune on new data is also important?

% 1. [some logic] T2I and its plugins are anywhere currently. The base model is trained on a larger dataset and the cost of retrain is high. When the new model is released, the plugins need to be trained and adapted individually, which is inefficient both in data and money XXX. UniPlug
Large text-to-image diffusion models~\cite{imagen, stable-diffusion, sdxl} have drawn the attention of both researchers and creators nowadays. 
Since these models are often trained on thousands of GPU days with millions of data pairs, the major development of the current research focuses on designing plug-and-play modules~\cite{animatediff, controlnet, t2i-adapter, lora}, which are commonly called plugins, to add new abilities on the pre-trained text-to-image models. 
People use plugins for photo creation~\cite{dreambooth,IP-Adapter,lora}, controllable drawing~\cite{controlnet,t2i-adapter}, and editing~\cite{SDEdit,fatezero}, both for image and video~\cite{animatediff,videocrafter1,show-1}.
The development speed of downstream plugins is faster than the release of the base model since it is easier to train and enables many more different features. But when a larger foundation model~(\eg, SDXL~\cite{sdxl}) is released, all the downstream plugins need to be retrained for this upgraded model, which takes much more time for maintenance and upgradation.

We aim to solve this inconvenient plugin incompatibility when upgradation by proposing a unified adapter network, where all the downstream plugins in the original base model~(\eg, Stable Diffusion v1.5~\cite{stable-diffusion}) can be directly used in upgraded model~(\eg, SDXL~\cite{sdxl}) via the proposed method. However, this task has a lot of difficulties.
First, when training different diffusion model versions, the compatibility of plugins is often not considered. Thus, the original connector of the plugin might not exist in the newly upgraded model due to dimension mismatch. Second, different plugins are applied in the different positions of the Stable Diffusion. For example, ControlNet~\cite{controlnet} and T2I-Adapter~\cite{t2i-adapter} are added at the encoder and decoder of the fixed denoising UNet respectively. LoRA~\cite{lora} are added after each linear layer of a fixed denoising UNet. This uncertainty makes it difficult to design a unified plugin.
% often comes with model structure changes, old plugins cannot be inserted into the upgraded model due to dimension mismatches 
Finally, although most current models are based on the latent diffusion model~\cite{stable-diffusion}, the latent space of each model is different. This gap is further boosted between the diffusion models in pixel and latent space.

% 2. we formulate this task a xxx , which has the challenge of A, B, C.

% 3. To solve this problem, we do D, E, F.
We propose X-Adapter to handle above difficulites. In detail, inspired by ControlNet~\cite{controlnet}, we consider X-Adapter as an additional controller of the upgraded model. To solve the problem of the connector and the position of different plugins, we keep a frozen copy of the base model in the X-Adapter. Besides, we design several mapping layers between the decoder of the upgraded model and X-Adapter for feature remapping. In training, we only train the mapping layers concerning the upgraded model without any plugins. Since the base model in the X-Adapter is fixed, the old plugins can be inserted into the frozen diffusion model copy in the X-Adapter. After training, we can sample two latent for X-Adapter and an upgraded model for inference. To further boost the performance, we also propose a two-stage inference pipeline by sequentially inference Stable Diffusion v1.5 first and then the SDXL inspired by SDEdit~\cite{SDEdit}. Experiments show that the proposed method can successfully upgrade the plugins for larger models without specific 
retraining. We also conduct numerical experiments to show the effectiveness of two widely used plugins, \ie, ControlNet~\cite{controlnet}, and LoRA~\cite{lora}. 

In summary, the contribution of this paper can be summarized as:

\begin{itemize}
    \item We target a new task in the large generative model era where we need to update plugins for different foundational models.
    \item We propose a general framework to enable upgraded model compatible with pretrained plugins. We propose a novel training strategy that utilizes two different latent with mapping layers. Besides, we design two kinds of inference strategies to further boost the performance.
    \item Experiments show the proposed methods can successfully make old plugins work on upgraded text-to-image model with better performance compared to the old foundational model.
\end{itemize}

% 4. Our contribution

% \xd{We aim to solve a very practical problem in the era of training and using Large Text-to-Image Generative Models~(\ie, Stable-Diffusion~\cite{sd}), where these base models have typically trained on the very large scale datasets~(\eg, LAION-2B~\cite{laion-2b}) with tones of GPU cost. Besides, these models typically contain various downstream plugin models~(ControlNet~\cite{controlnet}, T2i-Adaptor~\cite{t2i-adaptor}, LoRA~\cite{lora}) for custom and controllable usages. However, ,,,}
\section{Related Works}
\label{sec:formatting}

%-------------------------------------------------------------------------
% \subsection{T2I and its plugins}
\noindent\textbf{Diffusion Model for Text-to-Image Generation.}
Diffusion models are initially proposed by ~\citet{Diffusion}, and have recently been adapted for image synthesis~\cite{DiffusionModel, VariationalDiffusion}. Beyond unconditional image generation, the text-to-image diffusion models~\cite{stable-diffusion} is an important branch of the image diffusion model, since it leverages larger-scale datasets for training. In these networks,
% typically achieved by encoding text to latent features during generation. 
Glide~\cite{GLIDE} proposes a transformer~\cite{attention} based network structure. Imagen~\cite{imagen} further proposes a pixel-level cascaded diffusion model to generate high-quality images. Different from pixel-level diffusion, the technique of Latent Diffusion Models~(LDM)~\cite{stable-diffusion} conducts diffusion in a latent image space~\cite{VAE}, which largely reduces computational demands. Stable Diffusion v1.5~\cite{SD1.5} is a large-scale pre-trained latent diffusion model. Stable Diffusion v2.1~\cite{SD2.1} and SDXL~\cite{sdxl} are the following versions of Stable Diffusion v1.5 by optimizing latent space, network structure, and training data. Compared to Midjourney~\cite{Midjourney} and DALL~\cite{DALL-E2, DALL-E3}, SDXL achieves state-of-the-art results.

\noindent\textbf{Plugins for Text-to-Image Diffusion Model.}
Since the stable diffusion model~\cite{stable-diffusion} is open-sourced, plug-and-play modules, commonly referred to as ``plugins", significantly expand the capabilities of pre-trained text-to-image (T2I) models. GLIGEN~\cite{gligen} adds an additional gate attention for grounded generation. LoRA~\cite{lora} is a general parameter-efficient training method that allows us to fine-tune the stable diffusion for stylization and customization easily. Dreambooth~\cite{dreambooth} and Textual Inversion~\cite{textinversion, celebbasis} customize personal concepts by finetuning the pre-trained diffusion model. IP-Adapter~\cite{IP-Adapter} extends these works for universal image variation. Besides, ControlNet~\cite{controlnet} and T2I-Adapter~\cite{t2i-adapter} add spatial conditioning controls to diffusion models by incorporating an extra network to encode conditions. AnimateDiff~\cite{animatediff} allows a personalized T2I model to generate videos with high temporal consistency by adding a temporal module. Although these plugins are powerful, it is unfeasible to apply an old plugin to an upgraded T2I model, which significantly hampers the development and application of diffusion models. 

\noindent\textbf{Parameter-Efficient Transfer Learning.}
Our topic is also related to parameter-efficient transfer learning since we need to remedy the domain gap when upgrading.
The emergence of large-scale pre-trained models, \eg, Stable Diffuions~\cite{stable-diffusion}, CLIP~\cite{clip}, has highlighted the importance of the effective transfer of these foundational models to downstream tasks. Parameter-efficient Transfer Learning~(PETL) methods~\cite{pmlr, taca, revisit} add additional parameters to the original model to overcome the domain gaps between the pre-trained dataset and target tasks. PMLR~\cite{pmlr} introduces an adapter that consists of a down-sampling layer and an up-sampling layer and inserts it into Transformer~\cite{attention} blocks. ~\citet{revisit} bridge the domain gap by aligning the dataset's distribution. ~\citet{taca} propose a task-agnostic adapter among various upstream foundation models. Similar to upgrading the CLIP for visual understanding~\cite{taca}, our objective is to enable upgraded diffusion models compatible with all kinds of plugins. 

%TECA~?

% \clearpage

% \section{Model Upgrade for Diffusion Model}

\section{Methods}

% \xiaodong{COMMON: try to find reference of ``hidden state" in other papers, it should be named as ``latent"?}

% \xiaodong{Actually, our method aims to solve the problem of plugin compatibility, so the name of this section might not be suitable. the better section name is just ``method", with a subsection to target the core and preliminary of our work.}

\begin{figure}[t]
    \centering
    \includegraphics[width=\columnwidth]{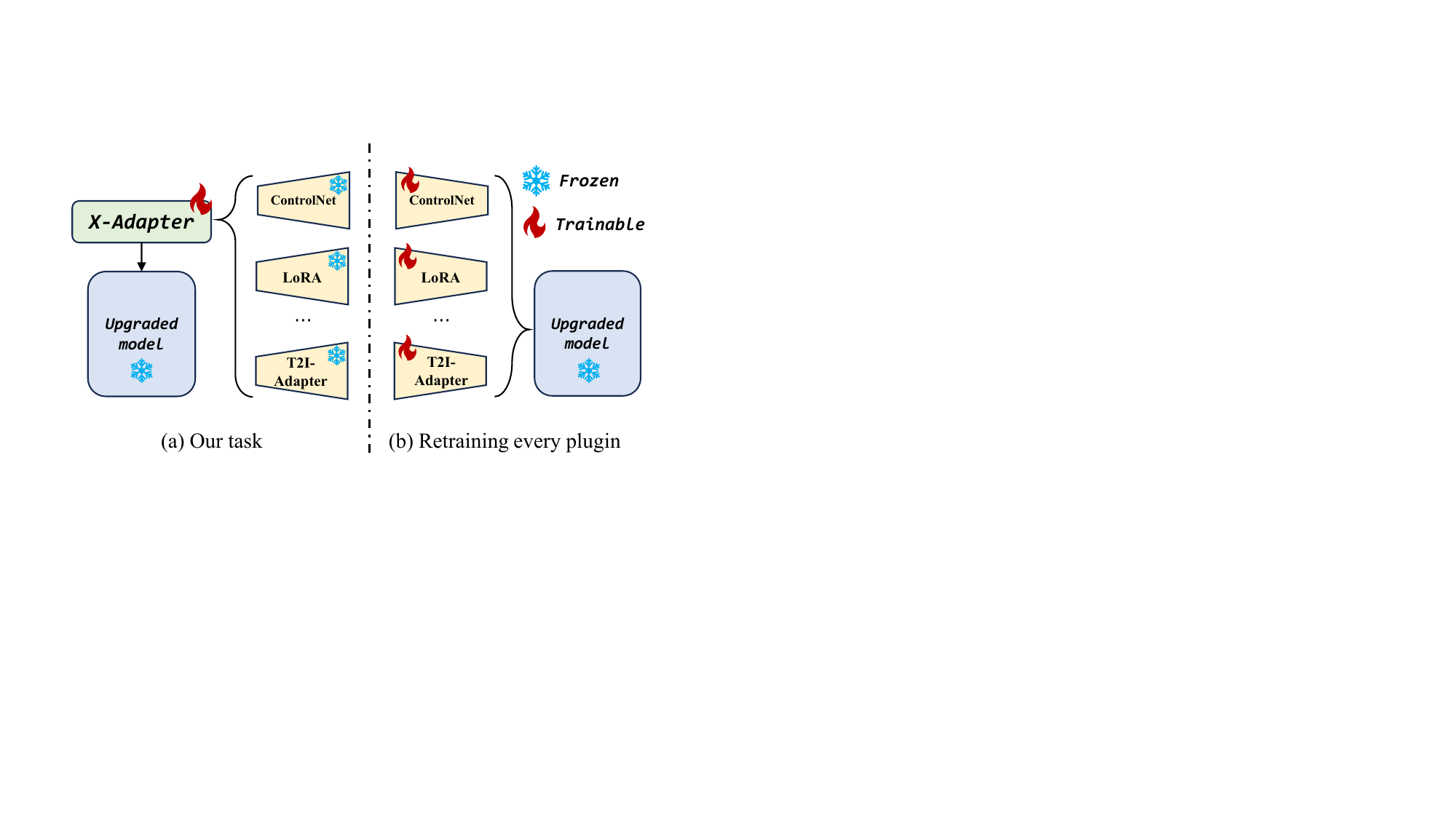}
    \vspace{-2em}
    \caption{\textit{\textbf{Task Definition.}} Different from the previous method to train each plugin individually, our method only trains a single X-Adapter to all the fixed downstream plugins. }
    \vspace{-1em}
    \label{fig:task_define}
\end{figure}

\subsection{Task Definition}

% 1. define our task
% 2. What we want to solve
% 3. The benefit of our task and potential solution 
% \xiaodong{More details about our task and show a preview figure.}
% \jw{should we first discuss the importance/motivation of the method here?}

We aim to design a universal compatible adapter~(X-Adapter) so that plugins of the base stable diffusion model can be directly utilized in the upgraded diffusion model.
As shown in Fig.~\ref{fig:task_define}, given a powerful pre-trained text-to-image diffusion model $M_{new}$~(\ie, SDXL~\cite{sdxl}), we aim to design a universal adapter X-Adapter so that all the pre-trained down-stream plugins~(\eg, ControlNet~\cite{controlnet}, T2I-Adapter~\cite{t2i-adapter}, LoRA~\cite{lora}) on $M_{base}$~(\ie, Stable Diffusion v1.5~\cite{stable-diffusion}) can work smoothly on $M_{new}$ without requiring additional training. Thanks to this universal adaption, we highlight some potential benefits:

\noindent\textit{(i) Universal Compatibility of Plugins from Base Model.} A naive idea to apply a plugin network to the new model is to directly train the specific downstream plugin individually. However, take ControlNet~\cite{controlnet} family as an example, it would require training more than ten different networks to achieve the original abilities. Differently, our method only needs to train \textit{one} version-to-version adapter in advance and enable direct integration of pre-trained plugins from the base model, \ie, Stable Diffusion v1.5~\cite{stable-diffusion}.

\noindent\textit{(ii) Performance Gain with respect to Base Model.} Since original plugins are only trained on the base model, their power is also restricted due to the limited generative capability. Differently, our adapter can improve the performance of these plugins by the upgraded models since these new models are typically more powerful in terms of visual quality and text-image alignments.

\noindent\textit{(iii) Plugin Remix Across Versions.} Since we retain the weights of both the base and upgraded models, our method also enables the use of plugins from both models~(\eg ControlNet of Stable Diffusion v1.5 and LoRA of SDXL can work together smoothly as if ControlNet were originally trained on SDXL). It largely expands the applicability of the plugins from different development stages of the text-to-image community.

% \noindent\textit{(iv) Ease the datasets' requirement to train higher-resolution plugins}. Train high-quality plugins on larger upgraded model~(\eg, SDXL~\cite{sdxl}) also needs high-quality visual datasets. Since we can directly upgrade all the plugin models, our method enjoys the high-quality generated output from the upgraded model directly.

\subsection{Preliminary: Latent Diffusion Model}
Before introducing our method, we first introduce the Latent Diffusion Model~(LDM~\cite{stable-diffusion}), since most of the open-source models are based on it. LDM extends denoising diffusion model~\cite{ddpm} for high-resolution image generation from text prompt, which first uses a VAE~\cite{VAE}'s encoder $\mathcal{E}$ to compress the RGB image $x$ into latent space $z$. 
% After that, a diffusion-based denoising UNet U  and a pre-trained CLIP~\cite{clip} text encoder $\mathcal{C}$ is used to generate the latent from pure Gaussian noise $\rho$ and text prompt $p$ with $T$ time stamp iteratively, and a decoder of auto-encoder is used to decode the latent $l$ to image space. For training, it can sample a single step $t$ to train the denoising UNet $\mathcal{U}$, which can be written as:
After that, a UNet~\cite{unet} $\epsilon_\theta$ is used to remove added noise from a noisy latent. Formally, $\epsilon_\theta$ is trained using the following objective:
% \begin{equation}
%     \arg_{min} || \mathcal{U}(l_{t-1}, \mathcal{C}(p), t)  - l_{t} ||^2,
% \end{equation}
\begin{equation}
\min _\theta E_{z_0, \epsilon \sim N(0, I), \bm{t} \sim \text { Uniform }(1, T)}\left\|\epsilon-\epsilon_\theta\left(\bm{z}_t, \bm{t}, \bm{c} \right)\right\|_2^2,
\end{equation}
% where $t$ is current step and $l_{0} = \rho$. \xiaodong{need fixed equation and make it consistent}
where $z_t$ is the noisy latent of $z$ from timestep $t$ and $c$ is the embedding of conditional text prompt.

\begin{figure*}[t]
    \centering
    \includegraphics[width= \textwidth]{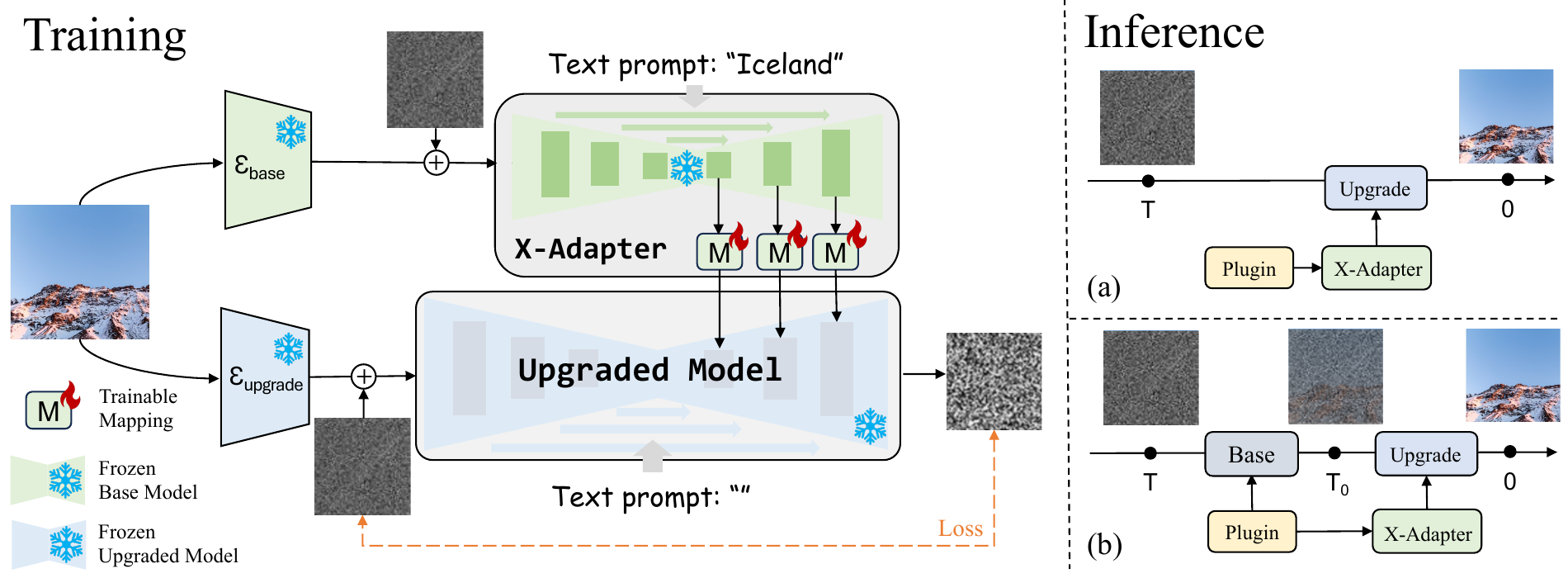}
    \vspace{-2em}
    \caption{\textit{\textbf{Method Overview}}. In training, we add different noises to both the upgraded model and X-Adapter under the latent domain of base and upgraded model. By setting the prompt of the upgraded model to empty and training the mapping layers, X-Adapter learns to guide the upgraded model. In testing, (a) we can directly apply the plugins on the X-Adapter for the upgraded model. (b) A two-stage influence scheme is introduced to improve image quality.}
    \vspace{-1em}
    \label{fig:main_framework}
\end{figure*}

\subsection{X-Adapter}
% \lm{
X-Adapter is built upon the base Stable Diffusion v1.5~\cite{stable-diffusion} to maintain the full support for the plugin's connectors. Additionally, in the decoder of each layer, we train an additional mapping network to map the features from the base model to the upgraded model~(\eg, SDXL~\cite{sdxl}) for guidance as shown in Fig.~\ref{fig:main_framework}.
In each mapper, a stack of three ResNet~\cite{ResNet} is utilized for dimension matching and feature extraction. 
Formally, suppose we have $N$ adapters and $\mathcal F_n(\cdot)$ denotes the $n^{th}$ trained mapper, given multi-scale feature maps $\bm{F}_{base} = \{\bm{F}_{base}^{1}, \bm{F}_{base}^{2}, ... ,\bm{F}_{base}^{N}\}$ from base model, guidance feature $\bm{F}_{mapper} = \{\bm{F}_{mapper}^{1}, \bm{F}_{mapper}^{2}, ... ,\bm{F}_{mapper}^{N}\}$ is formed by feeding $\bm{F}_{base}$ to the mapping layers. Note that the dimension of $\bm{F}_{mapper}$ is the same as that of certain intermediate features of upgraded decoder layers. $\bm{F}_{mapper}$ is then added with those layers. In summary, the guidance feature extraction and fusion can be defined as the following formulation:
% \begin{equation}
% \begin{align}
\begin{eqnarray}
\vspace{-2em}
\bm{F}_{mapper}^{n} &=& \mathcal{F}_n(\bm{F}_{base}^{n}) \\
\bm{F}_{up}^{n} &=& \bm{F}_{up}^{n} + \bm{F}_{mapper}^{n}, n\in\{1, 2, ..., N\},
\vspace{-2em}
\end{eqnarray}
where $\bm{F}_{up}^{n}$ denotes upgraded model's $n^{th}$ decoder layer to fuse guidance feature.

\noindent\textbf{Training Strategy.}
As shown in Fig.~\ref{fig:main_framework}, given an upgraded diffusion model, X-Adapter is firstly trained in a plugin-free manner on the upgraded diffusion model for text-to-image generation. Formally, given an input image $\mathcal{I}$, we first embed it to the latent spaces $z_0$ and $\overline{z}_0$ via base and upgraded autoencoder respectively. Then, we randomly sample a time step $t$ from $[0, T]$, adding noise to the latent space, and produce two noisy latent $z_t$ and $\overline{z}_t$ for denoising. 
% $z_t$ and $\overline{z}_t$ are initial latents for base and upgraded model. 
Given timestep $\bm{t}$, the prompt $\bm{c}_b$ of X-Adapter and upgraded model's prompt $\bm{c}_u$, X-Adapter is trained with the upgraded diffusion network $\epsilon_\theta$ to predict the added noise $\epsilon$ by:
\begin{equation}
E_{\bm{\overline{z}}_0, \epsilon, \bm{t}, \bm{c}_b, \bm{c}_u}\left\|\epsilon-\epsilon_\theta\left( \bm{z}_t, \bm{t}, \bm{c}_u, \mathcal{X}_{Adapter}(\bm{\overline{z}}_t, \bm{t}, \bm{c}_b) \right)\right\|_2^2.
\end{equation} 

In the training process, the objective of the above loss function is to determine the offsets between the X-Adapter and the upgraded space.  Inspired by previous task-compatibility plugins for additional control signal~\cite{controlnet, t2i-adapter} and video generation~\cite{animatediff}, we find that the key to task-agnostic adaptation is to fix the parameters of the trained diffusion UNet. 
Thus, we freeze the parameters in the base model during training, which ensures that old plugins can be seamlessly inserted. To avoid affecting the original high-quality feature space of the upgraded model, we also freeze its parameters similar to conditional control methods~\cite{controlnet, t2i-adapter}. All text prompts $\bm{c}_u$ are set to an empty string inspired by \cite{bridgeNet}. Thus, the upgraded model provides the average feature space with an empty prompt, while X-Adapter learns the offset given base feature space, guiding the native upgraded model. Although $\bm{c}_u$ is set to empty during training, our experiments show that we do not need to adhere this rule during inference and X-Adapter works well with any $\bm{c}_u$ after training. After training, the plugins can naturally be added to X-Adapter for their abilities.

\noindent\textbf{Inference Strategy.}
% \input{figs/inference}
% During training, we retain the original structure of the base model and freeze all parameters, allowing any type of plugin trained on base model, including ControlNet, LoRA, IP-Adapter~\cite{IP-Adapter}, \etc., to be seamlessly inserted into the base model like no upgraded model exists at inference stage. Besides, we observe that the plugin's insertion does not change base model's hidden states, ensuring $\mathcal{X}_{Adapter}$'s guidance ability with the involvement of plugin. To further improve generation quality and guidance capability, we design a prompt selection and latent initialization strategy specifically for the inference phase.
During training, two bypasses' latents are encoded from the same image, naturally aligning with each other. 
However, since the latent space of the two models is different, during the inference stage, if the initial latents for two bypasses are randomly sampled~(Fig.~\ref{fig:main_framework}~(a)), this leads to a lack of alignment, potentially causing conflicts that affect the plugin’s function and image quality. 
% \xd{Another problem of inference is the inconsistency between two latent.}

To tackle this issue, inspired by SDEdit~\cite{SDEdit}, we propose a two-stage inference strategy as shown in Fig.~\ref{fig:main_framework}~(b). Given total timestep $T$, at the first stage, we randomly sample an initial latent $z_T$ for X-Adapter and run with plugins in timestep $T_0$ where $T_0 = \alpha T, \alpha \in [0, 1]$. At timestep $T_0$, the base model's latent $z_{T_0}$ will be converted to upgraded model's latent $\overline{z}_{T_0}$ by:
\begin{equation}
\overline{z}_{T_0} = \mathcal{E}_{up}(\mathcal{D}_{base}(z_{T_0})),
\end{equation}
where $\mathcal{D}_{base}$ denotes the base model's decoder and $\mathcal{E}_{up}$ denotes the upgraded model's encoder. $\overline{z}_{T_0}$ and $z_{T_0}$ will be initial latents for two bypasses at the second stage where the plugin will guide the upgraded model's generation through X-Adapter. We observe that for most plugins our framework performs optimally when $T_0 = \frac{4}{5}T$, \ie, the base model run 20\% of the time step for warmup and then runs our X-Adapter in the rest of the inference time directly. We give detailed ablations on this two-stage inference in the experiments.

\if
To make the new diffusion model compatible with old plugins, the most straightforward approach would be training an adapter for each plugin. Then, the old plugin can function on the new version through the adapter. However, this approach requires additional training for each plugin. For the diffusion community with thousands of plugins uploaded by users, upgrading them through this method is costly and inefficient.

We, therefore, introduce a task of hot-plugging model upgrade[] for diffusion model. Old plugins can be directly applied to new diffusion model without additional training. To achieve this objective, the feature space of the new diffusion model must be compatible with the old version while remaining its advantage after integrating with old plugin.

{\bf Task objectives.} With the upgrade of diffusion model, we aim for an improvement in performance while retaining the functionality of plugins. Additionally, compatibility should be ensured across various types of plugins. Therefore, there are two main objectives for hot-plugging upgraded diffusion model.

(i) \emph {Performance improvement}: Since the new diffusion model always outperforms the old one in every aspect, the most crucial objective for hot-plugging model upgrade is to achieve performance improvement. Meanwhile, plugins retrained on the new diffusion model naturally work best with newer version, even though this incurs additional training costs. Formally, we denote the evaluation metric for generation quality as \emph {M($\cdot$)}, diffusion model as $\phi$ and plugin as $\psi$, the objective can be formulated as:
\begin{equation}
  M(\phi_{old}(\psi_{old})) < M(\phi_{new}(\psi_{old})) < M(\phi_{new}(\psi_{new}))
  \label{eq:important}
\end{equation}
Where $M(\phi_{old}(\psi_{old}))$ indicates the original performance of the system while $M(\phi_{new}(\psi_{new}))$ is the upper bound where the plugin is retrained on a new diffusion model. With the upgrade from $\phi_{old}$ to $\phi_{new}$,  our objective is to improve the overall performance without training $\psi_{old}$.

(ii) \emph {Universal compatibility}: Different plugins have distinct functionalities, and they are equally important for expanding the capabilities of diffusion model. Therefore, apart from performance gain, the second objective is to make new diffusion model compatible with different kinds of plugins. This necessitates that the new model can convert different attributes like semantics, style etc. from old plugins and, upon integration, does not harm its own feature space.

\fi
% \subsection{Network Structure} %Method Overview
% \jw{this section is a bit short. maybe can reorganize this sec with prev section on the Our method part}

\if

\fi

\section{Experiments}

% \xiaodong{should organize the experiments first, try to refine the below experiments one-by-one:}

% \xiaodong{1. comparison with other methods in detail methods:}

% \xiaodong{1.1 eg. lora, ControlNet, t2i-adaptor}

% \xiaodong{2. ablation study: }

% \xiaodong{2.1. the network structure?}

% \xiaodong{2.2. the importance of denoising strategy? }

% \xiaodong{2.3. ablation study on training dataset size? }

% \xiaodong{2.4. the importance of denoising strategy? }

% \xiaodong{2.4. hyper-parameters? }

% \xiaodong{is NULL text important?}

% \xiaodong{is frozen UNet in SD1.5 important for plugin?}

% \input{tables/baseline_comparison}

\subsection{Implementation Details}

\begin{table}[b]
\centering
\resizebox{\linewidth}{!}{
\begin{tabular}{@{}l@{\hspace{2mm}}c@{\hspace{2mm}}*{3}{c@{\hspace{2mm}}}r@{}}
% \begin{tabular}{@{}l@{\hspace{5mm}}c@{\hspace{1.5mm}}c@{\hspace{5mm}}c@{\hspace{4mm}}c@{\hspace{4mm}}c@{\hspace{3mm}}c@{\hspace{1mm}}c@{\hspace{1mm}}c@{\hspace{1mm}}c@{\hspace{1mm}}}

\toprule
Plugin: \textbf{ControlNet} & FID $\downarrow$ & CLIP-score $\uparrow$ & Cond. Recon. $\uparrow$\\
\midrule
SD 1.5~\cite{stable-diffusion} & 33.09 & 0.2426 & \textbf{0.33 ± 0.16}  \\
SDEdit~\cite{SDEdit} + SDXL & \textbf{30.86} & 0.2594 & 0.14 ± 0.10  \\
X-Adapter ~~+ SDXL & 30.95 & \textbf{0.2632} & 0.27 ± 0.13  \\
\midrule
\midrule
Plugin: \textbf{LoRA} & FID $\downarrow$ & CLIP-score $\uparrow$ & Style-Sim $\uparrow$\\
\midrule
SD 1.5~\cite{stable-diffusion} & 32.46 & 0.25 & -  \\
SDEdit~\cite{SDEdit} + SDXL & 30.11 & 0.2584 & 0.72  \\
X-Adapter ~~+ SDXL & \textbf{29.88} & \textbf{0.2640} & \textbf{0.83}  \\
\bottomrule
\end{tabular}
}
\vspace{-1em}
\caption{Quantitative evaluation against baselines. }
\label{table:Quantitative_baseline}
\vspace{-2em}
\end{table}

\begin{figure*}[t]
    \centering
    \includegraphics[width=1\textwidth]{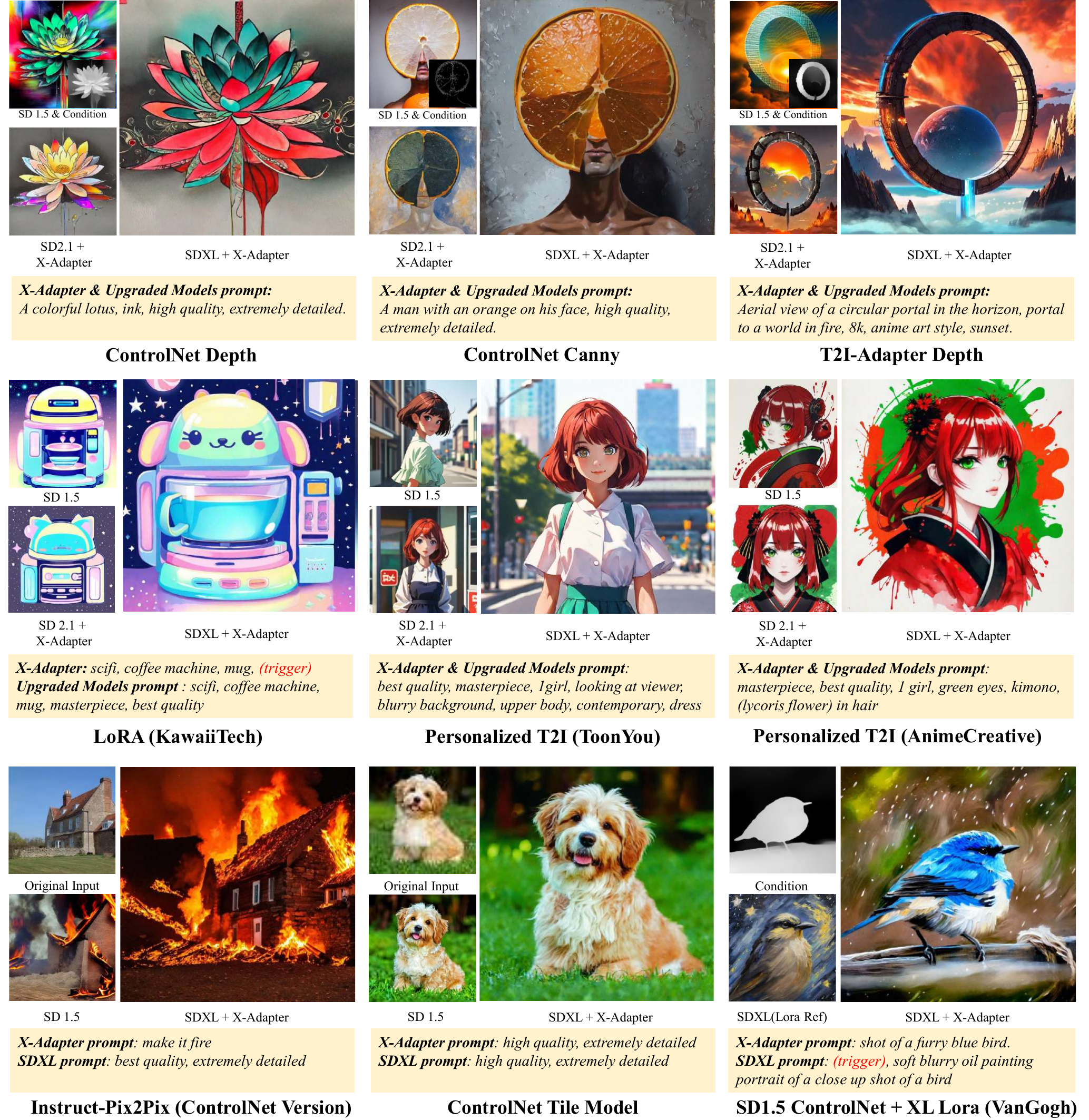}
    \vspace{-1.5em}
    \caption{\textbf{\textit{Qualitative Results on Different Plugins.}} The showcases of different results on SDXL and SD 2.1 based on the proposed X-Adapter and pre-rained SD 1.5 plugins. We show the corresponding prompts in the yellow box. }
    \vspace{-1em}
    \label{fig:qualitative_result}
\end{figure*}

We implement X-Adapter using Stable Diffusion v1.5~\cite{stable-diffusion} as the base model, and SDXL~\cite{sdxl} base as the main upgraded model. Mapping layers of X-Adapter are placed at the base model's last three decoder blocks. Notice that we also train our X-Adapter for Stable Diffusion v2.1~\cite{SD2.1}, which shows promising results shown as Fig.~\ref{fig:qualitative_result}.
% To upgrade the base model to Stable Diffusion XL, $\mathcal{X}_{Adapter}$'s output is inserted into the upgraded model's mid-block and first two decoder blocks. 
For training, we select a subset of Laion-high-resolution containing 300k images for X-Adapter training. In our experiments, the input image is resized into $1024\times1024$ for the upgraded model and $512\times512$ for the base model. We utilize the AdamW optimizer with a learning rate of $1e^{-5}$ and a batch size of 8. The model is trained for 2 epochs using 4 NVIDIA A100 GPUs.

% \subsection{Metrics}

% 
% 1. qualitative results
% 
% 2. quantitative results
% ControlNet,Lora => (1.5, 1.5+SDEdit+SDXL)
% 
% 3. user study

% \subsection{Compare with other methods}
% Controlnet, LoRA. SDXL, SD2.1

\begin{figure}[t]
    \centering
    \includegraphics[width=\columnwidth]{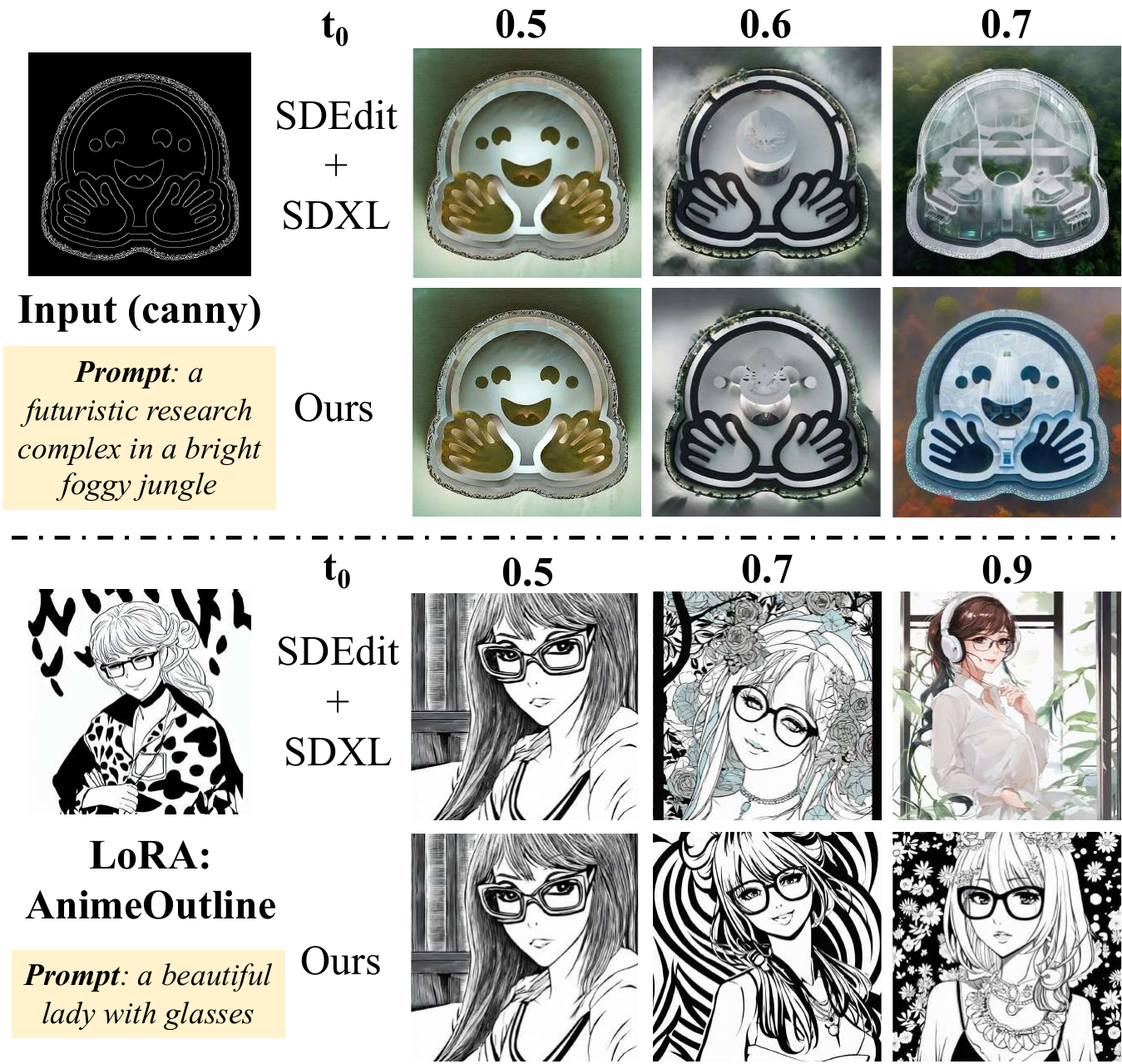}
    \vspace{-2em}
    \caption{\textit{\textbf{Visual Comparison to baseline under different $t_0$}}. We choose ControlNet~\cite{controlnet} and LoRA~\cite{lora} to evaluate different methods under semantic and style control. Specifically, we choose AnimeOutline\cite{animeoutline}, a LoRA specialized in black and white sketch generation. We sample three $t_0$ for each plugin. We observe that baseline loses style control (turn black and white to color) and semantic control as $t_0$ increases while our method maintain the controllability with the usage of X-Adapter.}
    \label{fig:baseline_comparison}
    \vspace{-1em}
\end{figure} 
\begin{figure}[t]
    \centering
    \includegraphics[width=\columnwidth]{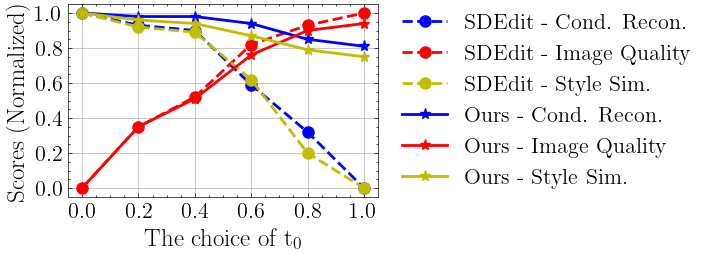}
    \vspace{-2em}
    \caption{\textit{\textbf{Quantitative evaluation under different $t_0$}}. Baseline loses style control and semantic control as $t0$ increases while our method preserves functionality of plugins}
    \label{fig:baseline_comparison_table}
    \vspace{-1em}
\end{figure} 

\subsection{Comparisons}
% Settings and Metrics?
% \noindent\textbf{User study.}
\noindent\textbf{Experiment setting.}
We choose two representative plugins~(ControlNet~\cite{controlnet} and LoRA~\cite{lora}), to evaluate the performance of the proposed method, since they represent two valuable applications of semantic and style control. We evaluate the performance gain our method achieves as well as plugin functionality retention. For ControlNet, we choose canny and depth to test our method under dense and sparse conditions. We utilize the COCO validation set, which contains 5,000 images, to evaluate each method. 
% For each image, our method and other methods will infer it once as our final result. 
For LoRA~\cite{lora}, We use AnimeOutline~\cite{animeoutline} and MoXin~\cite{moxin} to test the style control plugin. We select 20 prompts from civitai~\cite{civitai} for each LoRA, generating 50 images per prompt using random seeds. To eliminate SDXL~\cite{sdxl}'s effect on style, SDXL's prompt only focus on image's content, and X-Adapter's prompt will include LoRA's trigger words and style-related words.
As for evaluation metrics, we use Frechet Inception Distance~(FID) to measure the distribution distance over images generated by our method and original SDXL, which indicates image quality, as well as text-image clip scores. We also calculate the condition reconstruction score following ControlNet~\cite{controlnet} and style similarity following StyleAdapter~\cite{styleadapter} to evaluate the plugin's functionality. The style similarity is measured between the generation of our method and the base model. 

% \maketitle

% \noindent\textbf{User study.} Sample 20 controlnet and 20 lora, rank them in terms of "image quality", "fidelity of condition"(for controlnet), "fidelity of style"(for lora)

\noindent\textbf{Comparison to base model.}
We select Stable Diffusion v1.5~\cite{SD1.5} as our base model. The quantitative result is shown in Tab.~\ref{table:Quantitative_baseline}. It shows that our method achieves a balance between image quality and preservation of plugin's function.

\noindent\textbf{Comparison to baseline.} 
A naive approach is to consider SDXL as an editor for the output of the base Stable Diffusion v1.5 model, similar to SDEdit~\cite{SDEdit}. We select a timestep $t_0$, adding noise to the base model's generation to $t_0$ and denoising it using the upgraded model. 
We evaluate it under the same experiment setting as shown in Tab.\ref{table:Quantitative_baseline}. Note that the function of $t_0$ in SDEdit is similar to $T_0$ in our two-stage inference strategy. For both methods, the upgraded model is more influenced by the base model when $t_0$ is lower, obtaining more semantic features and style information from the base model, which leads to less optimal outcomes in terms of image quality. Conversely, a higher $t_0$ value decreases the base model's influence, leading to improved generation quality as shown in Fig.~\ref{fig:baseline_comparison}. This implies that the SDE-based method loses essential semantic details and style information (\ie, plugin's control) when $t_0$ is large, indicative of higher image quality. Conversely, X-adapter can maintain these controls and preserve the capabilities of the plugins even with a high $t_0$, ensuring high-quality generation with faithful plugin fidelity. To highlight the advantage of our method, we sampled six $t_0$ values at equal intervals between $[0,1]$ and conducted experiments on our method and baseline under these $t_0$. Fig.~\ref{fig:baseline_comparison} and Fig.~\ref{fig:baseline_comparison_table} illustrate the performance of different methods. 
We observe that although our method shows similar visual quality compared to the baseline, it better preserves the functionality of plugins. 
% \xiaodong{need to specify the value of $t_0$} 

% \paragraph{User Study}
% \xiaodong{add user study}

\begin{table}[t]
\centering
\resizebox{\linewidth}{!}{
\begin{tabular}{@{}l@{\hspace{2mm}}c@{\hspace{2mm}}*{3}{c@{\hspace{2mm}}}r@{}}
% \begin{tabular}{@{}l@{\hspace{5mm}}c@{\hspace{1.5mm}}c@{\hspace{5mm}}c@{\hspace{4mm}}c@{\hspace{4mm}}c@{\hspace{3mm}}c@{\hspace{1mm}}c@{\hspace{1mm}}c@{\hspace{1mm}}c@{\hspace{1mm}}}
\toprule
Plugin: \textbf{ControlNet} & Result Quality $\uparrow$ & Condition Fidelity $\uparrow$ \\
\midrule
SD 1.5~\cite{stable-diffusion} & 3.23 ± 0.12 & \textbf{4.21 ± 0.32} \\
SDEdit~\cite{SDEdit} + SDXL & 4.14 ± 0.57 & 2.46 ± 0.17 \\
X-Adapter ~~+ SDXL & \textbf{4.46 ± 0.43} & 3.92 ± 0.26 \\
\midrule
\midrule
Plugin: \textbf{LoRA} & Result Quality $\uparrow$ & Style Fidelity $\uparrow$\\
\midrule
SD 1.5~\cite{stable-diffusion} & 2.93 ± 0.09 & - \\
SDEdit~\cite{SDEdit} + SDXL & 3.92 ± 0.53 & 3.45 ± 0.33 \\
X-Adapter ~~+ SDXL & \textbf{4.38 ± 0.25} & \textbf{4.14 ± 0.29}  \\
\bottomrule
\end{tabular}
}
\vspace{-1em}
\caption{\textit{\textbf{User Study}}. We report the user preference ranking (1 to 5 indicates worst to best) of different methods. }
\label{table:User_Study}
\vspace{-1.5em}
\end{table}

\noindent\textbf{User study.} 
Users evaluate the generation results of our method with ControlNet~\cite{controlnet} and Lora~\cite{lora}. For ControlNet, we collect 10 canny conditions and depth conditions, then assign each condition to 3 methods: Stable Diffusion v1.5, SDEdit + SDXL, and X-Adapter. We invite 5 users to rank these generations in terms of \textit{“image quality”} and \textit{"fidelity of conditions"}. For LoRA, we collect 10 prompts and also assign them to these three methods. Users rank these generations in terms of \textit{“image quality”} and \textit{"style similarity"}. We use the Average Human Ranking (AHR) as a preference metric where users rank each result on a scale of 1 to 5 (lower is worse). The average rankings are shown in Tab~\ref{table:User_Study}.

\subsection{Qualitative Results on Multiple Plugins}
% \xiaodong{Add discussion}
As shown in Fig.~\ref{fig:qualitative_result}, we show the qualitative results of the proposed X-Adapter on both SD 2.1 and SDXL in various pretrained plugins on Stable Diffusion v1.5 to show the advantages. 
We present representative examples of conditional generation~(ControlNet Depth, ControlNet Canny, T2I-Adapter Depth), the personalization style~(LoRA Model~\cite{KawaiiTech}, Personalized Model~\cite{toonyou, AnimeCreative}) and the Image Editing Methods~(ControlNet-based InstructPix2Pix and ControlNet Tile). Finally, we show the plugin remix in our methods, where the plugins~\cite{VanGoghPortraiture} in SDXL can also directly cooperate with the Stable Diffusion v1.5 plugin~(\eg, ControlNet in our case).

\subsection{Ablative Study}
\begin{figure}[t]
    \centering
    \includegraphics[width=\columnwidth]{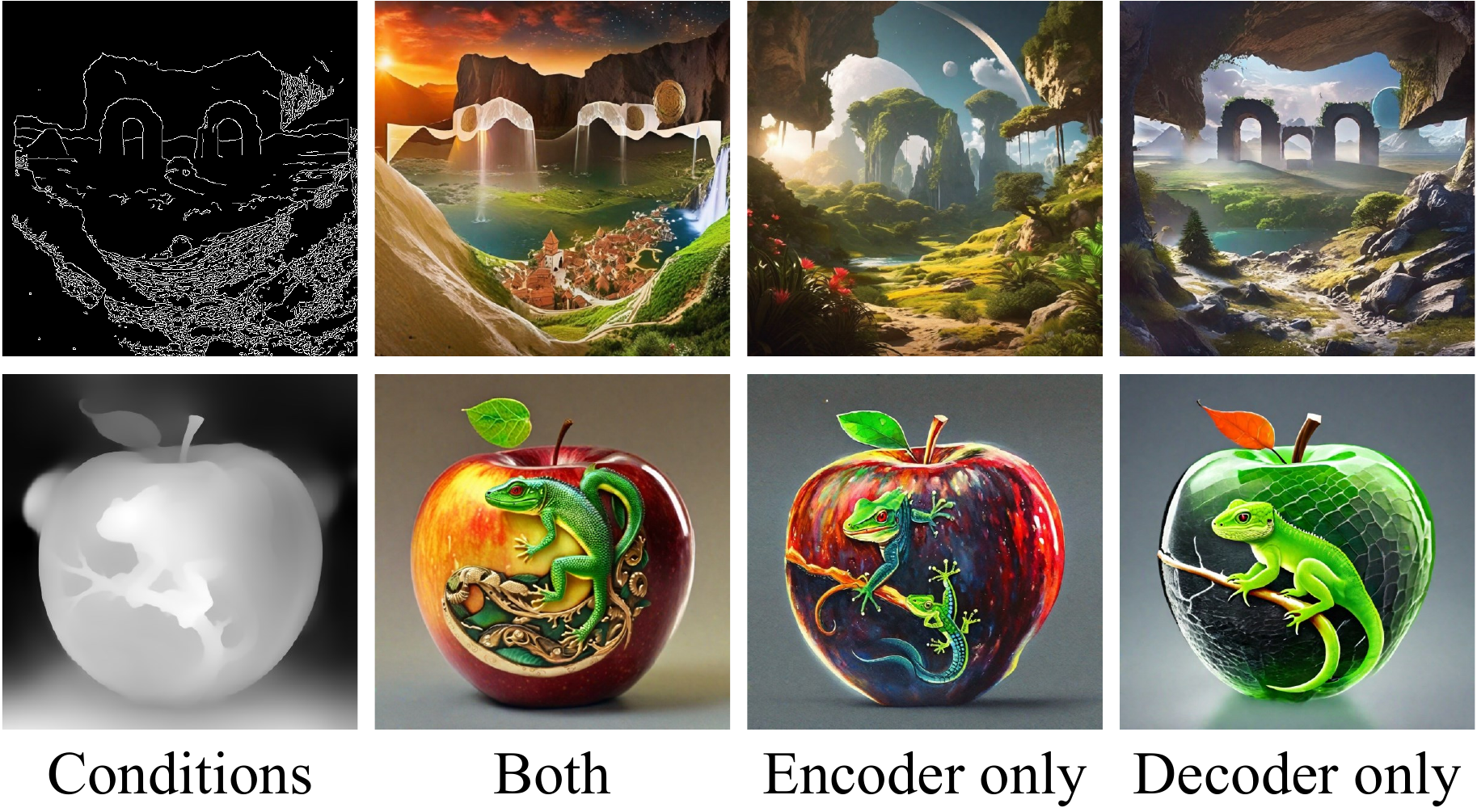}
    \vspace{-2.em}
    \caption{\textit{\textbf{Ablation of module to insert mapping layers}}. The key to better guidance ability is to retain encoder's feature space. Prompts: \textit{“a fantastic landscape~/~an apple with a lizard in it”}. 
}
    \vspace{-1.2em}
    \label{fig:Module selection}
\end{figure}

\noindent\textbf{Where to insert mapping layer?}
We study the effect of inserting mapping layers into different modules: (1) Encoder; (2) Decoder; (3) Both encoder and decoder. Fig.~\ref{fig:Module selection} indicates that the decoder-only strategy shows the strongest guidance capability since it does not harm the encoder's feature space and only performs guidance during generation. See also the supplementary material for quantitative results for different module selection.
% Tab.~\ref{table:ablation_module} indicates that the decoder-only strategy shows the strongest guidance capability since it does not harm the encoder's feature space and only performs guidance during generation. We also show different setting's results in Fig.~\ref{fig:Module selection}. Encoder-only setting lacks of guidance ability while the encoder-and-decoder setting disrupted the upgraded model's feature space, causing a decline in image quality. 
% \xiaodong{need to explain fig7}

\begin{figure}[t]
    \centering
    \includegraphics[width=\columnwidth]{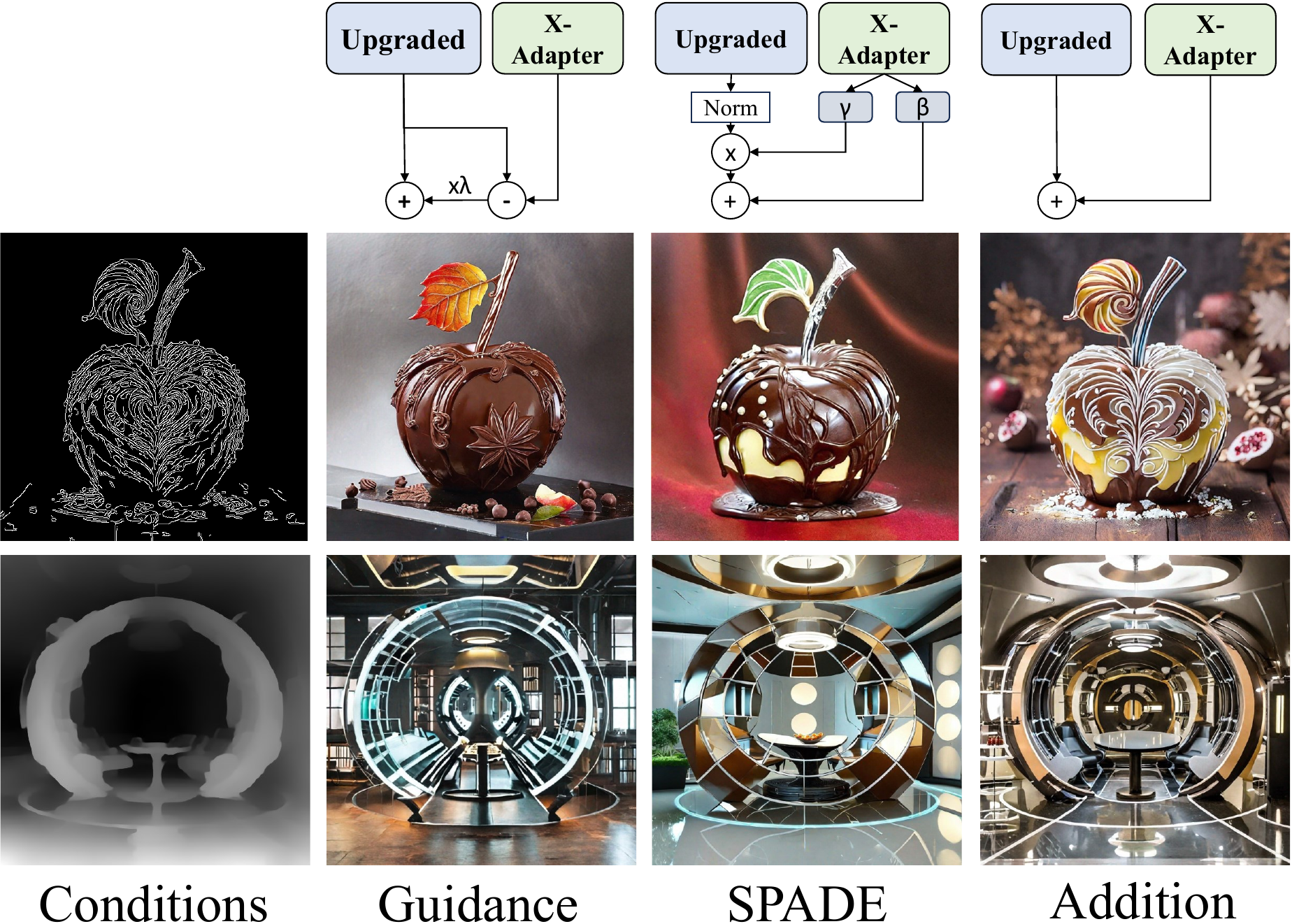}
    \vspace{-1.5em}
    \caption{ \textit{\textbf{Ablation of different fusion types}}. The result shows that fusing features through addition can maximize the restoration of the condition. The text prompts are: \textit{``A chocolate apple"} and \textit{``A research room"}. 
}
    \vspace{-2.00em}
    \label{fig:FusionType}
    
\end{figure}

\noindent\textbf{How do mapping layers guide the upgraded model?}
We explored three methods for integrating guidance features into the upgraded model. Given guidance feature $a$ and upgraded model's feature $b$, new feature $c$ is formed by (1) addition fusion: $c=a+b$ (2) guidance fusion: $c=b+\lambda(a-b)$ where $\lambda$ can be adjusted by users (3) SPADE:  
$c=\gamma(a)(norm(b)) + \beta(a)$ where $\gamma$ and $\beta$ are two networks following SPADE~\cite{SPADE}'s design. 
Fig.~\ref{fig:FusionType} presents a visual comparison of different ablation fusion types. We find that addition is the most effective way to provide guidance for the upgraded model.

\begin{figure}[t]
    \centering
    \includegraphics[width=\columnwidth]{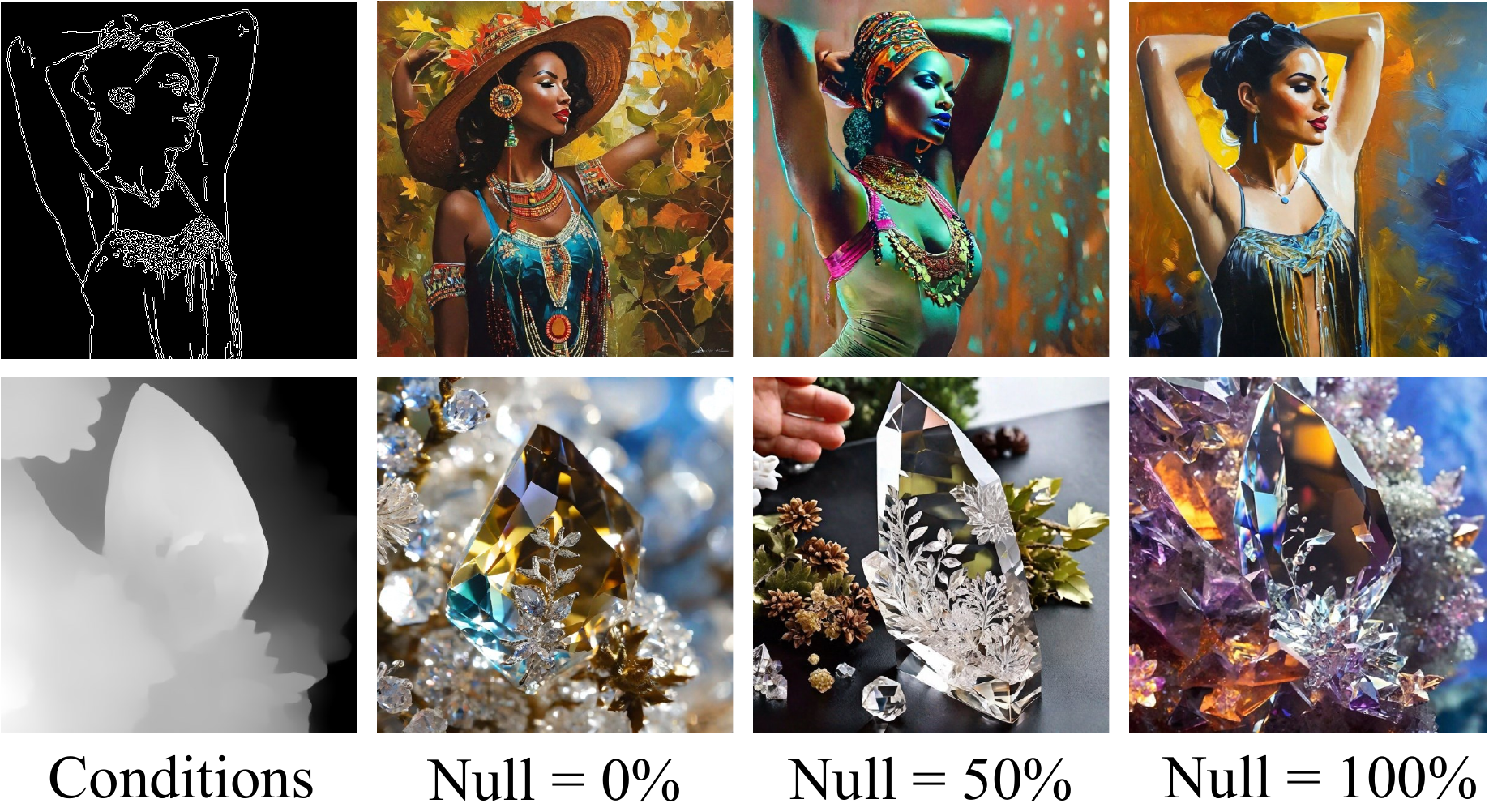}
    \vspace{-2em}
    \caption{\textit{\textbf{Ablation of different null probability during training}}. Increasing the percentages of null text prompts in the upgraded model can enhance ${X}_{Adapter}$'s guidance ability. Text prompts are: \textit{``A painting of a beautiful woman"} and \textit{``A world of crystal"} from top to bottom. 
    }
    \vspace{-1.5em}
    \label{fig:NullProb}
\end{figure}

\noindent\textbf{Is using empty text important in the upgraded model?}
To demonstrate the effectiveness of the null-text training strategy, we train three models under 100\%, 50\%, and 0\% null probability. Fig.~\ref{fig:NullProb} indicates that reducing the capability of the upgraded model during training can maximize the guidance effect of X-Adapter.

\begin{figure}[t]
    \centering
    \includegraphics[width=\columnwidth]{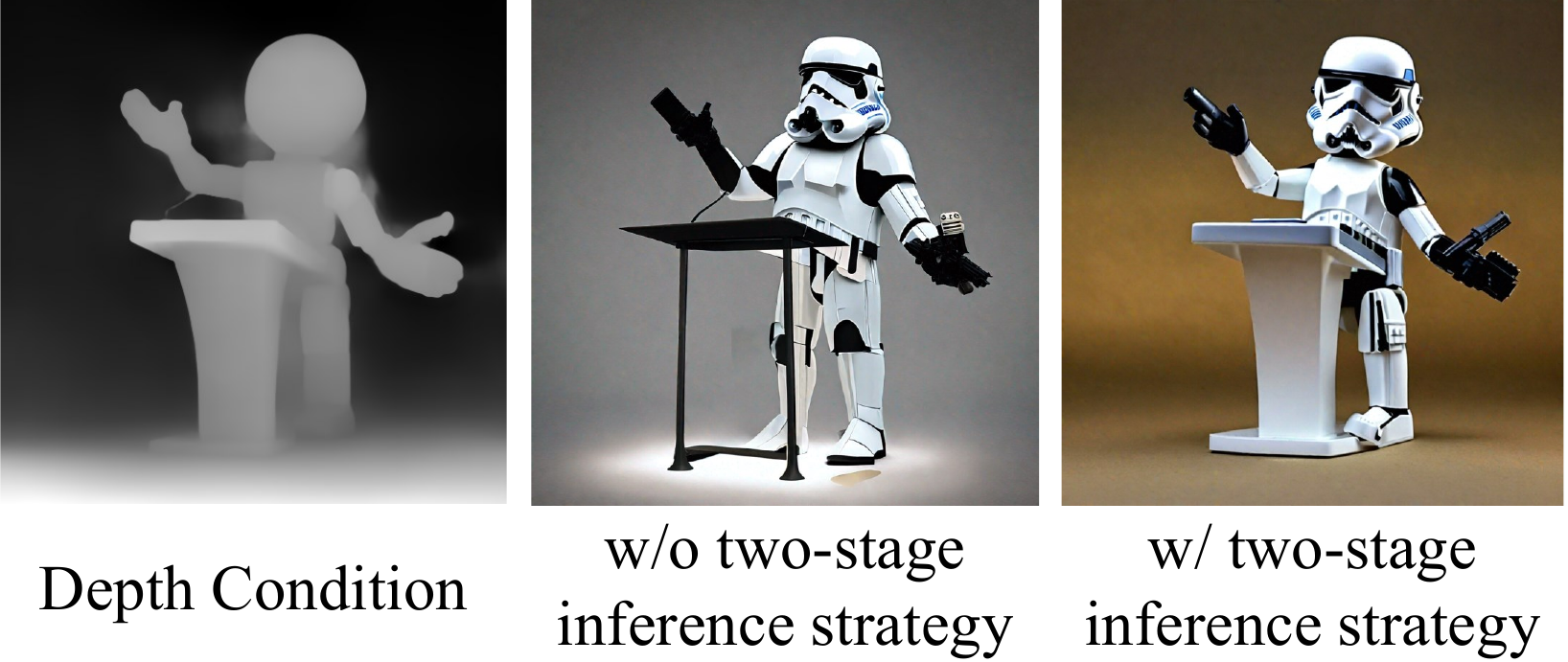}
    \vspace{-2.em}
    \caption{\textit{\textbf{Ablation of inference strategy.}} The result shows that X-Adapter can roughly reconstruct the condition even w/o the two-stage inference, and the two-stage inference has a better similarity. Text prompt: \textit{``stormtrooper lecture, photorealistic"}}
    \vspace{-2.em}
    \label{fig:InferenceAblation}
\end{figure}

\noindent\textbf{Is two-stage inference important?}
We study the effect of a two-stage denoising strategy by randomly sampling initial latents for X-Adatper and upgraded model. Our method still works effectively  without initial latents alignment as shown in Fig.~\ref{fig:InferenceAblation}. Adopting two-stage sampling strategy in inference further boosts performance in terms of conditional accuracy. 

\subsection{Discussion}
\begin{figure}[t]
    \centering
    \includegraphics[width=\columnwidth]{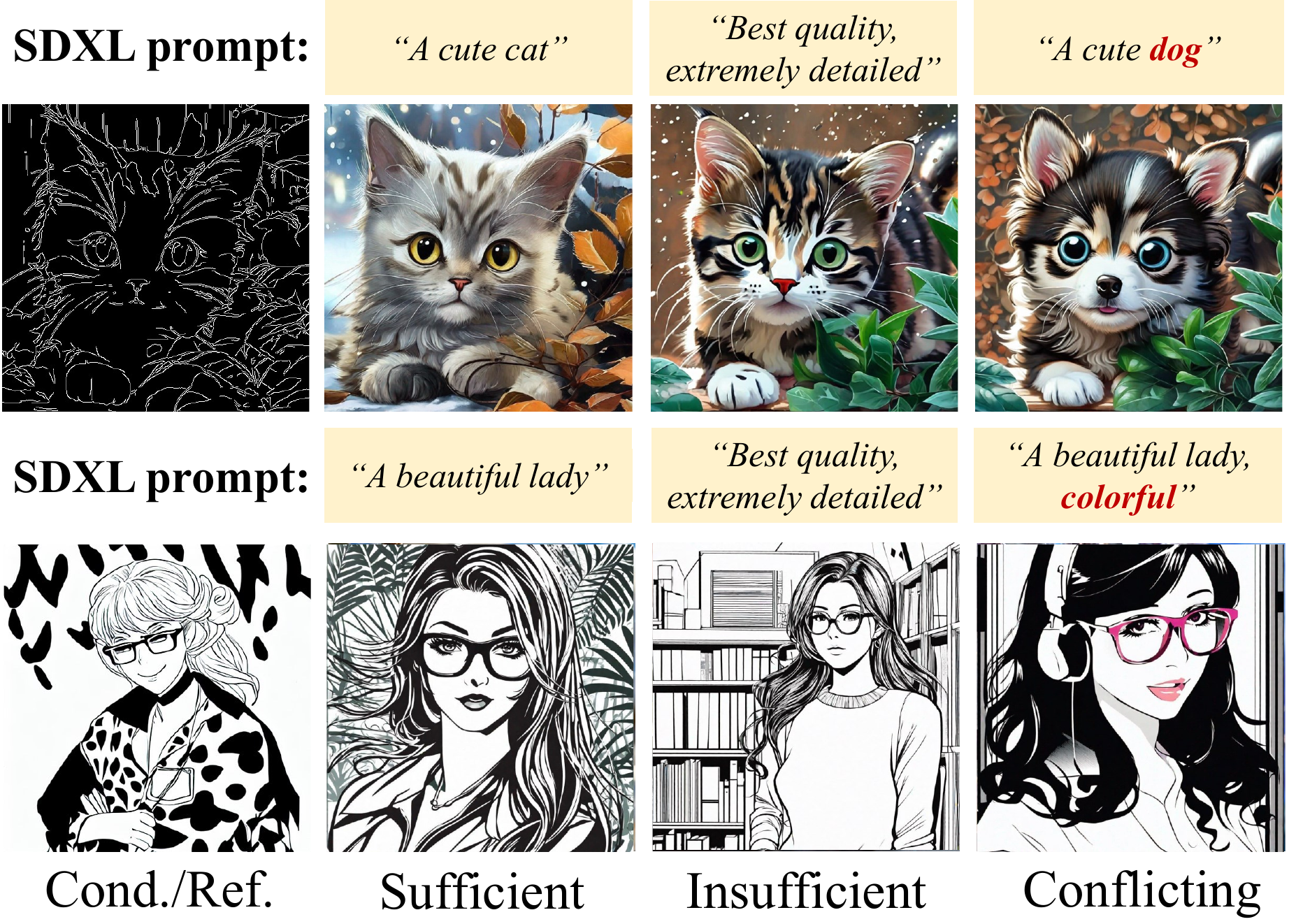}
    \vspace{-2em}
    \caption{\textit{\textbf{Prompt setting}}. Our method can still ensure the overall layout and style consistency even in case of prompt conflict.  LoRA\cite{lora} used here isAnimeOutline\cite{animeoutline}, an expert in black and white sketch generation. X-Adapter's text prompts are: \textit{``A cute cat"} and \textit{``A beautiful lady, (trigger words)"} from top to bottom. 
    }
    \vspace{-2em}
    \label{fig:PromptSetting}
\end{figure}
\noindent\textbf{Prompt Setting.}
We always set clear and sufficient prompts for X-Adapter, therefore we study three different prompt settings of SDXL: (1) Sufficient prompts which are semantically consistent with X-Adapter's prompts (2) Insufficient prompts. The default insufficient prompt in this paper is "best quality, extremely detailed". (3) Conflicting prompts which change the meaning of X-Adapter's prompts. Fig.~\ref{fig:PromptSetting} shows that our method can still maintain overall layout and style consistency even in case of prompt conflict.

\noindent\textbf{Plugin Remix.}
Our method naturally supports plugins from both X-Adapter~(\eg SD1.5~\cite{SD1.5}) and upgraded model~(\eg SDXL~\cite{sdxl}) since we retain all connectors by freezing parameters of these two models. The bottom right picture of Fig.~\ref{fig:qualitative_result} shows a combination of Stable Diffusion v1.5's ControlNet and SDXL's LoRA, generating results that follow LoRA's style and condition's semantics. It indicates that our method can bridge community resources across different diffusion model versions~(\eg SD1.5, SD2.1~\cite{SD2.1}, SDXL).

\noindent\textbf{Limitation.}
Although our method achieves impressive results, it still has some limitations. For some plugins to generate personalized concepts, \eg, IP-Adapter~\cite{IP-Adapter}, our method might not maintain the identity well. We give examples in the supplementary material for visualization. This is because the custom plugins work on the text-encoder other than the feature space concepts that are not directly injected into the upgraded model other than fused as guidance. 
% How to manipulate the upgraded model at a more abstract level is our next research direction.
Since our method has already made some universal plugin upgrades, we leave the capability of the concept customization as future work. 
% \input{figs/fail_case}

% \vspace{-1em}
\section{Conclusion}
% \vspace{-1em}
In this paper, we target a new task of upgrading all the downstream plugins trained on old diffusion model to the upgraded ones. To this end, we propose X-Adapter, 
% which comprises a copied network structure and weights of the base model, 
which comprises a copied base model 
% and a series of mapping layers between two decoders for feature mapping. 
and mapping layers between two decoders for feature mapping. During training, we freeze the upgraded model and set its text prompt to empty text to maximize the function of X-Adapter. In testing, we propose a two-stage inference strategy to further enhance performance. We conduct comprehensive experiments to demonstrate the advantages of the proposed methods.
% in terms of compatibility and visual quality.
% \vspace{-1em}
\section{Acknowledge}
% This research is supported by National Research Foundation, Singapore and A*STAR, under its RIE2020 Industry Alignment Fund – Industry Collaboration Projects (IAF-ICP) grant call (Grant No. I2001E0059) – SIA-NUS Digital Aviation Corp Lab.
This research is supported by National Research Foundation, Singapore and A*STAR, under its RIE2020 Industry Alignment Fund – Industry Collaboration Projects (IAF-ICP) grant call (Grant No. I2001E0059) – SIA-NUS Digital Aviation Corp Lab. Mike Zheng Shou is supported by the Ministry of Education, Singapore, under the Academic Research Fund Tier 1 (FY2022) Award 22-5406-A0001.

{
    \small
    \bibliographystyle{ieeenat_fullname}
    \bibliography{main}
}


\end{document}

\clearpage
\setcounter{page}{1}
\setcounter{figure}{0}
\setcounter{table}{0}
\maketitlesupplementary
% \title{Supplementary Material}
\appendix

\section{Ablation on module selection}
We provide quantitative results of ablation on module to insert mapping layers. Tab.~\ref{table:ablation_module} indicates that the decoder-only strategy shows the strongest guidance capability since it does not harm the encoder's feature space and only performs guidance during generation. 

\section{Limitation}
\begin{figure}[H]
    \centering
    \includegraphics[width=\columnwidth]{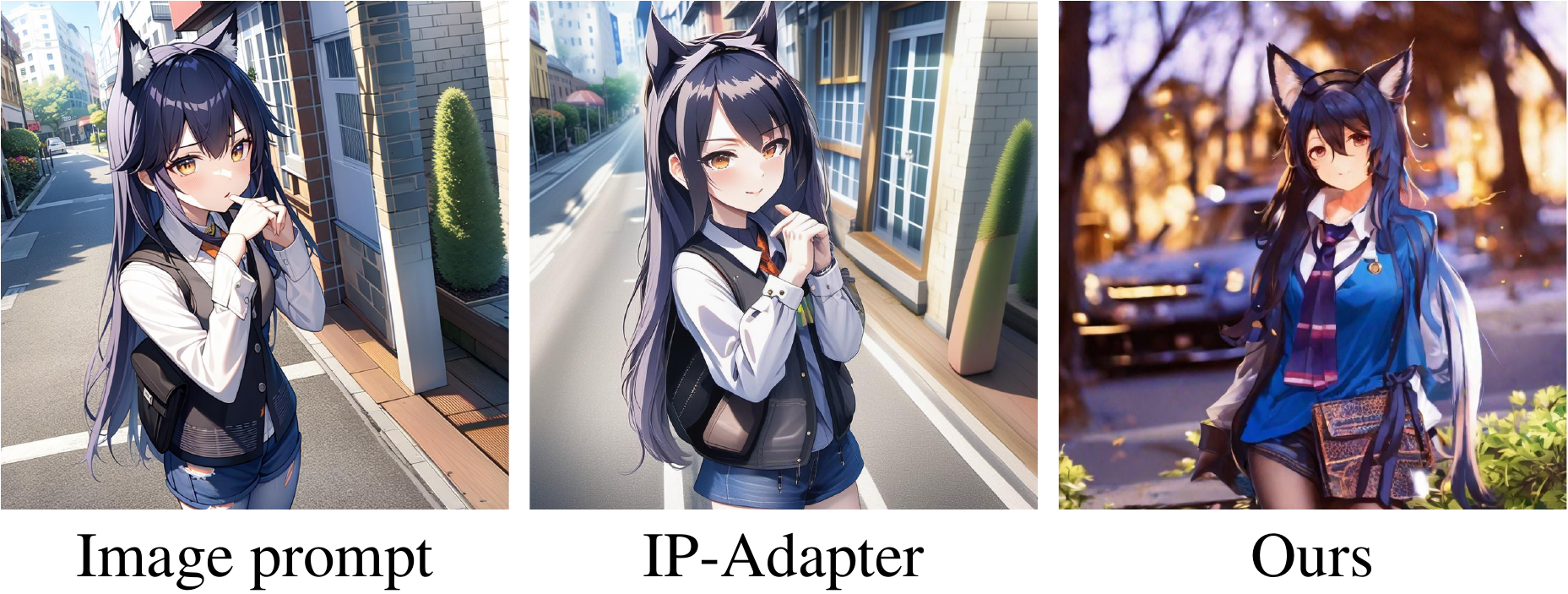}
    \vspace{-2em}
    \caption{\textit{\textbf{Limitations}}. In IP-Adapter~\cite{IP-Adapter}, although our method can produce relatively identity-consistent results, the details, \eg, clothes, are still different from the original model.}
    \label{fig:fail_case}
\end{figure}

For plugins to generate personalized concepts, \eg, IP-Adapter~\cite{IP-Adapter}, our method might not maintain the identity well like shown in Fig.~\ref{fig:fail_case}. This is because the custom plugins work on the text-encoder other than the feature space concepts that are not directly injected into the upgraded model other than fused as guidance. 
\section{Qualitative result}
\begin{figure*}[t]
    \centering
    \includegraphics[width=1\textwidth]{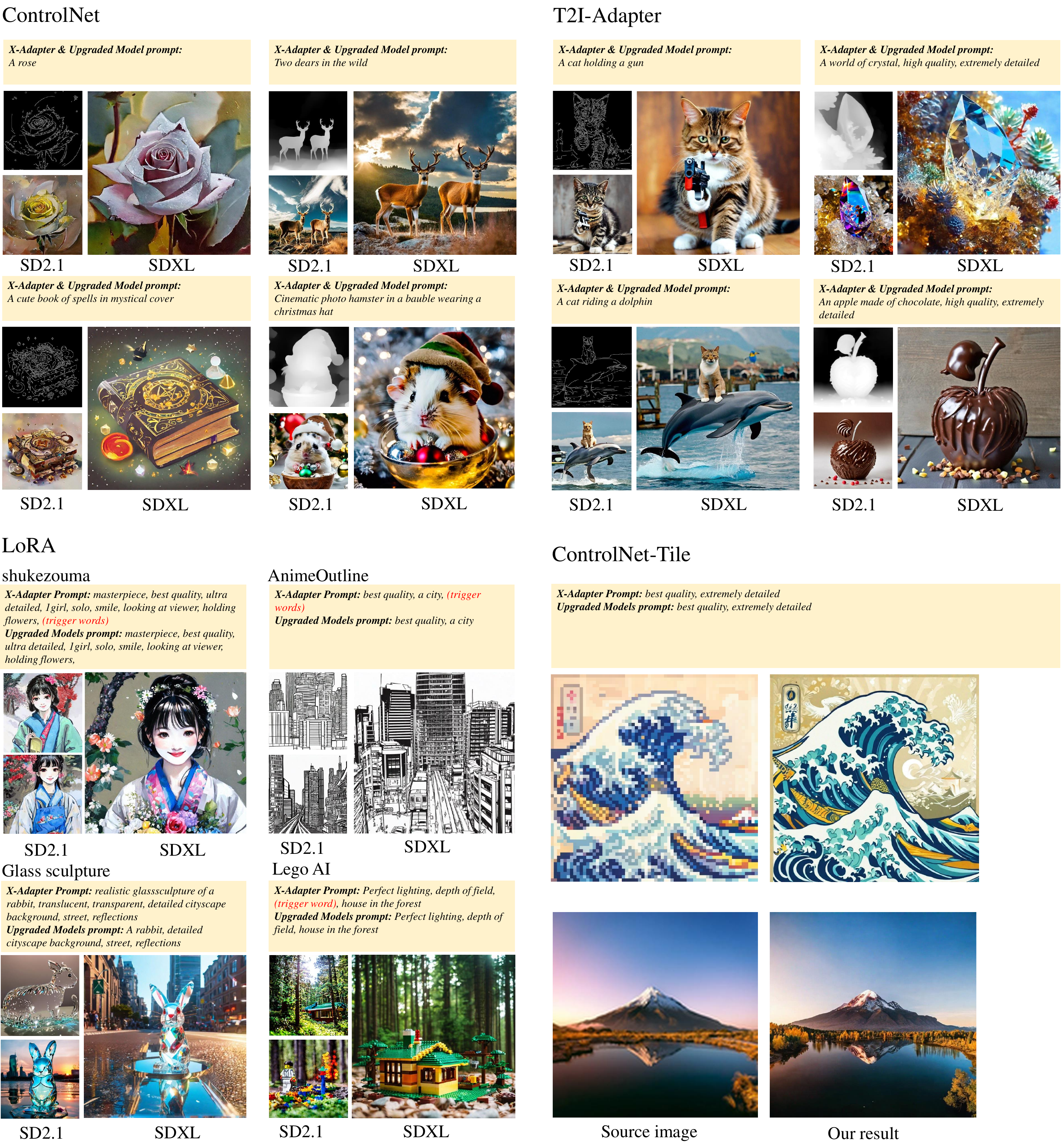}
    \vspace{-1.5em}
    \caption{\textbf{\textit{Qualitative Results on Different Plugins.}}}
    \label{fig:supp_qualitative_result}
\end{figure*}

We provide more qualitative result on Controlnet~\cite{controlnet}, T2I-Adapter~\cite{t2i-adapter} and LoRA~\cite{lora} as shown in Fig.~\ref{fig:supp_qualitative_result}

% \begin{table}[t]
\begin{table}[H]

\centering
\resizebox{\linewidth}{!}{
\begin{tabular}{@{}l@{\hspace{2mm}}c@{\hspace{2mm}}*{3}{c@{\hspace{2mm}}}r@{}}
% \begin{tabular}{@{}l@{\hspace{5mm}}c@{\hspace{1.5mm}}c@{\hspace{5mm}}c@{\hspace{4mm}}c@{\hspace{4mm}}c@{\hspace{3mm}}c@{\hspace{1mm}}c@{\hspace{1mm}}c@{\hspace{1mm}}c@{\hspace{1mm}}}

\toprule
Plugin: \textbf{ControlNet} & FID $\downarrow$ & CLIP-score $\uparrow$ & Cond. Recon. $\uparrow$\\
\midrule
Encoder \& Decoder & 38.37 & 0.26 & 0.24 ± 0.15  \\
Encoder only & 37.32 & 0.26 & 0.23 ± 0.14  \\
Decoder only & \textbf{30.95} & \textbf{0.26} & \textbf{0.27 ± 0.13} \\
\midrule
\midrule
Plugin: \textbf{LoRA} & FID $\downarrow$ & CLIP-score $\uparrow$ & Style-Sim $\uparrow$\\
\midrule
Encoder \& Decoder & 36.71 & 0.26 & 0.79  \\
Encoder only & 35.54 & 0.26 & 0.80  \\
Decoder only & \textbf{29.88} & \textbf{0.26} & \textbf{0.83}  \\
\bottomrule
\end{tabular}
}
\vspace{-1em}
\caption{\textit{\textbf{Ablation on module to insert mapping layers.}} For encoder only and decoder only, we use the same mapping layers to map the frozen base model to the upgraded model's encoder and decoder separately. For both the encoder and decoder, we use two identical mapping layers and insert them into the encoder and decoder.}
\label{table:ablation_module}
% \vspace{-1em}
\end{table}

% \begin{figure}[t]

% \maketitle

% \section{Rationale}
% \label{sec:rationale}
% % 
% Having the supplementary compiled together with the main paper means that:
% % 
% \begin{itemize}
% \item The supplementary can back-reference sections of the main paper, for example, we can refer to \cref{sec:intro};
% \item The main paper can forward reference sub-sections within the supplementary explicitly (e.g. referring to a particular experiment); 
% \item When submitted to arXiv, the supplementary will already included at the end of the paper.
% \end{itemize}
% % 
% To split the supplementary pages from the main paper, you can use \href{https://support.apple.com/en-ca/guide/preview/prvw11793/mac#:~:text=Delete%20a%20page%20from%20a,or%20choose%20Edit%20%3E%20Delete).}{Preview (on macOS)}, \href{https://www.adobe.com/acrobat/how-to/delete-pages-from-pdf.html#:~:text=Choose%20%E2%80%9CTools%E2%80%9D%20%3E%20%E2%80%9COrganize,or%20pages%20from%20the%20file.}{Adobe Acrobat} (on all OSs), as well as \href{https://superuser.com/questions/517986/is-it-possible-to-delete-some-pages-of-a-pdf-document}{command line tools}.